%% file: TMM26.tex
\documentclass[lettersize,journal]{IEEEtran}
\usepackage{amsmath,amsfonts}
\usepackage{algorithmic}
\usepackage{algorithm}
\usepackage{array}
\usepackage[caption=false,font=normalsize,labelfont=sf,textfont=sf]{subfig}
\usepackage{textcomp}
\usepackage{stfloats}
\usepackage{url}
\usepackage{verbatim}
\usepackage{graphicx}
\usepackage{cite}
\usepackage{wrapfig}
\usepackage[dvipsnames]{xcolor}
\usepackage[table]{xcolor}
\usepackage{multirow}
\usepackage{bbding}
\usepackage{url}
\usepackage{amssymb}
\usepackage{mathtools}
\usepackage{amsthm}
\usepackage{microtype}
\usepackage{graphicx}
\usepackage{subcaption}
\usepackage{booktabs}
\usepackage{array}

\begin{document}

\title{EvolvingAgent: Curriculum Self-evolving Agent with Continual World Model for Long-Horizon Tasks}

\author{Tongtong Feng, Xin Wang,~\IEEEmembership{Member,~IEEE,} Zekai Zhou, Ren Wang, Yu-Wei Zhan, Guangyao Li, Qing Li, \\Wenwu Zhu,~\IEEEmembership{Fellow,~IEEE}

\thanks{Tongtong Feng, Xin Wang, Ren Wang, Yu-Wei Zhan, Guangyao Li, and Wenwu Zhu are with the Department of Computer Science and Technology, Beijing National Research Center for Information Science and Technology, Tsinghua University, Beijing 100084, China (E-mail: \{fengtongtong, xin{\_}wang, rwang2xx, zhanyw, guangyaoli, wwzhu\}@tsinghua.edu.cn). Zekai Zhou is with the Department of Computer Science, University of Sydney, Sydney, Australia (E-mail: zhouodywork@gmail.com). Qing Li is with the Department of Electronic Engineering, Tsinghua University, Beijing 100084, China (E-mail: soleilor@tsinghua.edu.cn). }
\thanks{Corresponding authors: Xin Wang and Wenwu
Zhu.}}

\markboth{Journal of \LaTeX\ Class Files,~Vol.~14, No.~8, August~2021}%
{Shell \MakeLowercase{\textit{et al.}}: A Sample Article Using IEEEtran.cls for IEEE Journals}


\maketitle

\begin{abstract}
Completing Long-Horizon (LH) tasks in open-ended worlds is an important yet difficult problem for embodied agents. Existing approaches suffer from two key challenges: (1) they heavily rely on experiences obtained from human-created data or curricula, failing to autonomously update and select multimodal experiences, and (2) they may encounter catastrophic forgetting issues when faced with new tasks, failing to autonomously update world knowledge. To solve these challenges, this paper presents {\bf EvolvingAgent}, a curriculum self-evolving agent with a continual World Model (WM), which can autonomously complete various LH tasks across environments through self-planning, self-control, and self-reflection, without human intervention. Specifically, EvolvingAgent contains three modules, i.e., i) the experience-driven task planner, which uses an LLM along with multimodal experiences to convert LH tasks into executable sub-tasks; ii) the WM-guided action controller, which leverages WM to generate low-level actions and incorporates a self-verification mechanism to update multimodal experiences; iii) the Curriculum Learning (CL) -based reflector, which implements a two-stage CL algorithm to select multimodal experiences for task-adaptive WM updates. By building a planner-controller-reflector closed-loop dynamic, the continual WM for EvolvingAgent can autonomously update multimodal experiences and world knowledge. We conducted extensive experiments on Minecraft, compared with existing methods, EvolvingAgent can improve 111.74{\%} average success rate, reduce more than 6x ineffective actions, and generalize to the Atari environment with human-level performance.
\end{abstract}


\section{Introduction}
\IEEEPARstart{L}{ong}-Horizon (LH) tasks \cite{feng2025u2udata, shen2025detach} are complex, multi-step tasks that require sustained planning, sequential decision-making, and extended execution over a prolonged period to achieve a final goal. These tasks are challenging, often exhibiting reward sparsity \cite{hafner2025mastering} and procedural diversity \cite{kwa2025measuring}. Completing LH tasks in open-ended worlds is an important yet difficult problem for embodied agents, such as logistics robots, surgical robots, and rescue robots. 

\begin{figure}[!t]
\centering
	\includegraphics[width=0.49\textwidth]{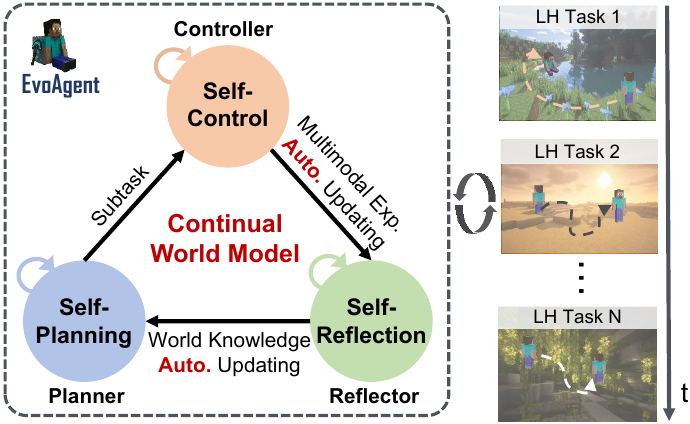}
	\centering
	\caption{The illustration of EvolvingAgent. EvolvingAgent can autonomously complete various long-horizon tasks across environments through self-planning, self-control, and self-reflection, without human intervention. The continual World Model (WM) for EvolvingAgent builds planner-controller-reflector closed-loop dynamic, which can autonomously update multimodal experiences and world knowledge.}
	\label{fig_1}
\end{figure} 

On the one hand, existing agents have made remarkable progress by utilizing expert data and domain-specific curricula created by humans, developing policies through Reinforcement Learning (RL) \cite{ren2025surfer}, Imitation Learning (IL) \cite{seo2025legato}, and Large Language Models (LLMs) \cite{Li_2025_CVPR}. On the other hand, recent studies \cite{kwa2025measuring} demonstrate that humans' ability to accomplish LH tasks in an open world relies on autonomous multimodal experience accumulation and world knowledge updates. In essence, autonomous world knowledge update serves as a meta-cognitive driver that not only guides action selection under partial observability but also enables context-aware adaptation to environmental dynamics, thereby resolving the local optimality issue inherent in LH task completion. 

{\it Completing LH tasks in open-ended worlds requires agents to achieve autonomous multimodal experience accumulation and world knowledge updates, like a baby thrives.}

Nevertheless, existing methods are hard to complete various LH tasks across environments from scratch: {\it 1) Failing to autonomously update and select multimodal experiences.} Most embodied agents assume that all training data are available from the beginning (such as IL-based or LLMs-based agents), which heavily rely on human-created data or curricula \cite{Li_2025_CVPR}. However, this assumption is unrealistic, as agents may encounter novel tasks or environments after deployment \cite{ Zhang2024VLABenchAL}. {\it 2) Failing to autonomously update world knowledge.} On the one hand, existing methods use LLMs (such as Voyager \cite{wang2023voyager}, Jarvis-1 \cite{Wang2023JARVIS1OM}) to represent world knowledge based on sampling historical experiences or use a graph (such as Optimus-1 \cite{li2024optimus1hybridmultimodalmemory}) to sparsely represent world knowledge, which requires human intervention and is hard to autonomously update. On the other hand, existing methods face catastrophic forgetting, where they lose previously obtained knowledge \cite{nayakLLaMARLongHorizonPlanning2025, hafner2025mastering} for learning new tasks, which are hard to autonomously update and transfer world knowledge for LH tasks across environments.

To solve these challenges, in this paper, we propose {\bf EvolvingAgent} (as shown in Fig. \ref{fig_1}), a curriculum self-evolving agent with a continual World Model (WM), which can autonomously complete various LH tasks across environments through self-planning, self-control, and self-reflection, without human intervention. EvolvingAgent contains three modules: i) The experience-driven task planner, which uses an LLM along with multimodal experiences, to incorporate self-state into the planning phase and convert LH tasks into executable sub-tasks; ii) The WM-guided action controller, which leverages WM to generate low-level actions and incorporates a self-verification mechanism to update multimodal experiences. iii) The Curriculum Learning (CL) -based reflector, which implements a two-stage CL algorithm to select experiences for task-adaptive WM updates. By using a model-based online RL setup and building a planner-controller-reflector closed-loop dynamic, the continual WM for EvolvingAgent can autonomously update multimodal experiences and world knowledge, filtering out invalid explorations and mitigating historical forgetting. 

We evaluate EvolvingAgent's performance in Minecraft \cite{fan2022minedojo}, a popular open-world environment. Extensive experiments demonstrate EvolvingAgent’s superiority: compared with existing methods, EvolvingAgent can achieve an average success rate improvement of 111.74{\%} and reduce ineffective actions by more than 6x. We also evaluate the generalization of EvolvingAgent in the Atari environment \cite{bellemare2013arcade}; EvolvingAgent outperforms existing methods and achieves human-level performance in some tasks. The contributions are summarized as follows:
\begin{itemize}
	\item We propose EvolvingAgent, a curriculum self-evolving agent, which can autonomously complete various LH tasks across environments through self-planning, self-control, and self-reflection, without human intervention.
	\item We design a novel continual WM for EvolvingAgent, building planner-controller-reflector closed-loop dynamic to autonomously update multimodal experiences and world knowledge.
	\item We conduct extensive experiments on Minecraft. EvolvingAgent can achieve an average success rate improvement of 111.74{\%}, reduce ineffective actions by more than 6x, and generalize to the Atari environment with human-level performance.
\end{itemize}

\section{Related Works}
\subsection{Embodied Long-horizon Tasks}
Embodied Long-Horizon (LH) tasks \cite{zhou2026gentle, shen2025detach, feng2025u2udata} refer to complex, multi-step tasks that require sustained planning, sequential decision-making, and extended execution over a prolonged period to achieve a final goal. Existing work on embodied agents completing LH tasks can be divided into two categories. One is Model-Based Reinforcement Learning (MBRL) \cite{medany2025model}. Embodied agents leverage MBRL to tackle LH tasks by interacting with environments and learning predictive world dynamics \cite{krinner2025accelerating}. The other is vision-language model-based (VLM) planning \cite{Roger2025RobinAS, hu2025vision}. Embodied agents leverage VLMs to decompose LH tasks into hierarchical sub-goals \cite{Liu2024ReLEPAN}, dynamically refine plans via memory-augmented reasoning \cite{lou2025explorevlm}, and align semantic intent with executable actions through iterative simulation \cite{yang2025guiding}, such as EmbodiedGPT \cite{EmbodiedGPT}, which bridges high-level planning with low-level control. However, they assume perfect knowledge of environments, rely on oracle feedback, and assume perfect execution of low-level policies, which makes it hard to adapt various LH tasks across environments in open worlds \cite{Zhang2024VLABenchAL}. 

\subsection{World Models}
World Models (WMs) empower embodied AI by building internal representations and making future predictions of the external world~\cite{feng2025embodied, ha2018world}. They serve as simulators of real environments that predict the future outcome of certain actions, and policies can be derived from them. Current research focuses on two paradigms: understanding the world through latent state representations \cite{assran2025v, hassan2025gem} and predicting future dynamics for planning and control \cite{zuo2025gaussianworld}. Recurrent State-Space Model (RSSM) \cite{ hafner2025mastering} is a classic world model structure. Representative example usages of them in MBRL include action searching~\cite{nayakLLaMARLongHorizonPlanning2025}, policy optimization within such simulators, or a combination of both~\cite{hafner2025mastering}. However, WMs currently struggle to prevent catastrophic forgetting \cite{Mattes2023HierosHI} due to their inability to maintain stable representations of previously learned environmental dynamics while adapting to new tasks, often exacerbated by shared parameter updates prior to knowledge \cite{Sun2024LearningLD}. 

\subsection{Curriculum Learning}
Curriculum Learning (CL) not only is a easy-to-hard heuristic, but also is a policy of difficulty modeling, schedule adaptation, and task-dependent deployment. In multimedia settings, this shift is particularly visible. For visual grounding, CLIP-VG~\cite{Xiao2024CLIPVG} shows that a self-paced curriculum can stabilize adaptation by gradually exploiting pseudo-language supervision rather than trusting noisy alignments from the beginning. For video understanding, exploring clip order in a self-supervised curriculum provides a structured way to organize temporal learning signals, improving downstream video applications beyond random pretraining order~\cite{Xiao2021VideoClipOrder}. In human pose estimation, DMH-CL~\cite{Dai2024DMHCL} further suggests that curriculum design can be tied to model hardness, allowing the training process to move according to the learner’s state rather than a fixed schedule. A related idea appears in data-free knowledge distillation, where category-aware curricula are used to control the transfer path from easier categories to harder ones~\cite{Li2024CategoryAwareKD}. Recent work has started to systematize the field through multimodal weak supervision~\cite{Mai2022WSMCL}, unified benchmarking~\cite{Zhou2024CurBench}, and psychologically grounded difficulty estimation and pacing strategies~\cite{Meng2025PUDF}. These studies suggest that modern curriculum learning is becoming less a handcrafted ordering trick and more a trainable mechanism for matching data exposure to model maturity. However, CL is currently not applied to solving embodied long-horizon tasks, nor has it achieved autonomous experience updating.

\begin{figure*}[!t]
\centering
	\includegraphics[width=\textwidth]{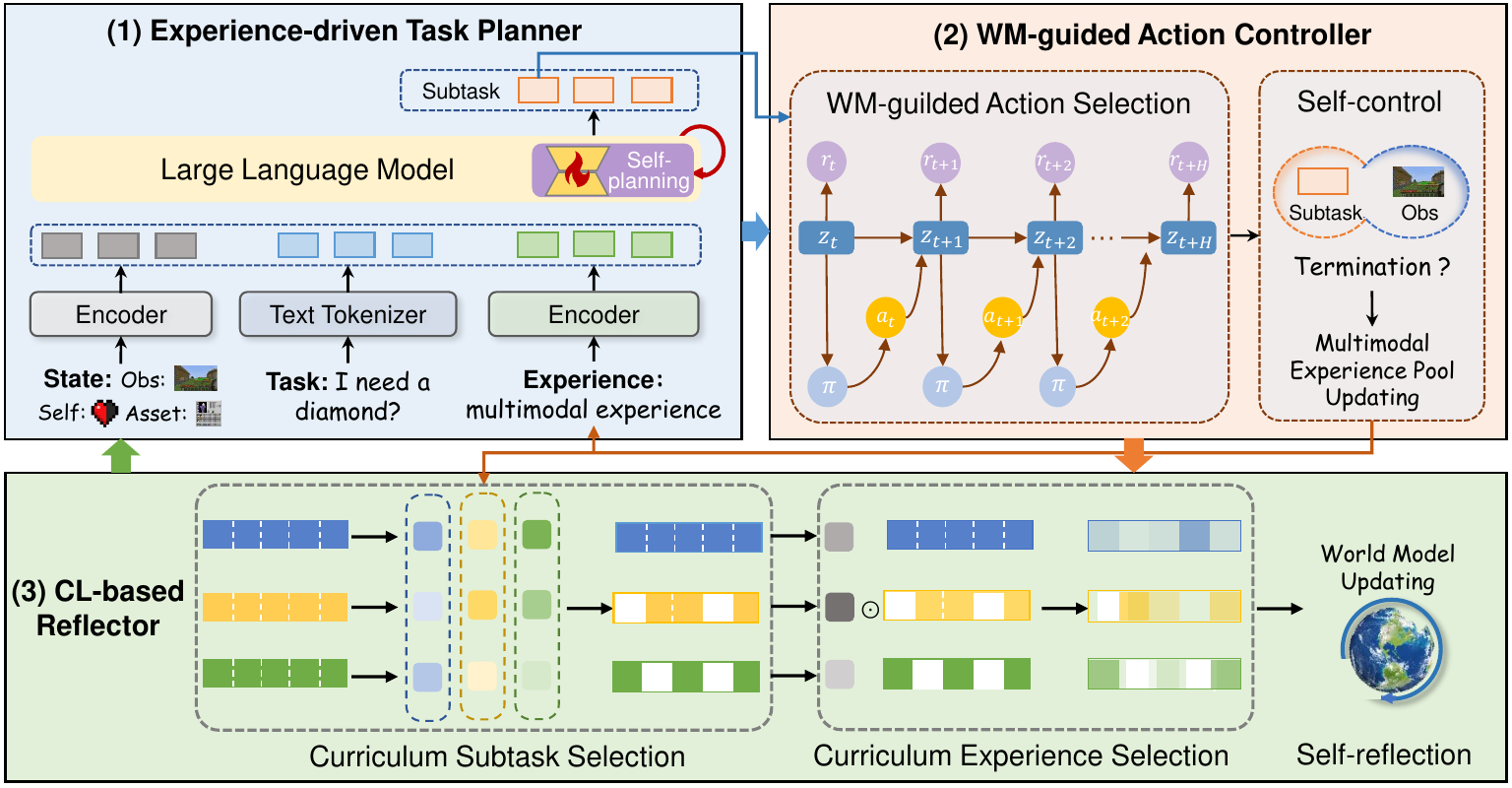}
	\centering
	\caption{The overview of EvolvingAgent, which includes an experience-driven task planner, a WM-guided action controller and a Curriculum Learning (CL) -based reflector. The continual WM for EvolvingAgent, building planner-controller-reflector closed-loop dynamic, can autonomously update multimodal experiences and world knowledge.}
	\label{fig_2}
\end{figure*}

\section{Preliminaries}
\subsection{Online Model-based Reinforcement Learning}
Reinforcement Learning (RL) is typically formulated as a Markov Decision Process (MDP) defined by the tuple $(\mathcal{S}, \mathcal{A}, P, R, \gamma)$, where $\mathcal{S}$ is the state space, $\mathcal{A}$ is the action space, $P(s'|s,a)$ is the transition dynamics, $R(s,a)$ is the reward function, and $\gamma \in [0,1)$ is the discount factor. The goal is to learn a policy $\pi(a|s)$ that maximizes the expected cumulative reward:
\begin{equation}
J(\pi) = \mathbb{E}_{\pi, P}\left[\sum_{t=0}^\infty \gamma^t R(s_t, a_t)\right],
\end{equation}
In Model-based Reinforcement Learning (MBRL), the agent explicitly learns a model $\mathcal{M}$, which includes an approximate dynamics model $\hat{P}_\theta(s'|s,a)$ and a reward model $\hat{R}_\phi(s,a)$, parameterized by $\theta$ and $\phi$, respectively. These models are trained to minimize empirical prediction errors over observed transitions $\mathcal{D} = \{(s_i, a_i, s'_i, r_i)\}$: 
\begin{equation}
\mathcal{L}_{\text{model}}(\theta, \phi) = \mathbb{E}_{(s,a,s',r)\sim\mathcal{D}}\left[\|s' - \hat{P}_\theta(s,a)\|^2 + \|r - \hat{R}_\phi(s,a)\|^2\right],
\end{equation}
Using the learned models, the agent performs planning to optimize its policy. For example, in value iteration, the state-value function $V(s)$ is iteratively updated via the Bellman equation:
\begin{equation}
V(s) \leftarrow \max_a [\hat{R}_\phi(s,a) + \gamma \mathbb{E}_{s'\sim\hat{P}_\theta(\cdot|s,a)} V(s')].
\end{equation}
In online MBRL, an agent interacts with the environment iteratively for $K$ rounds with the goal of learning a sequence to minimize $\mathcal{L}_{\text{model}}(\theta, \phi)$.

\subsection{Recurrent State-Space Model}
Recurrent State-Space Model (RSSM) \cite{ hafner2025mastering, hafner2019planet} is a classic world model structure, which can predict latent states and rewards from high-dimensional observations. RSSM contains 6 modules. 1) Encoder, maps observation \( o_t \) to a stochastic latent state \( s_t = (h_t, z_t) \), where \( h_t \) is a deterministic RNN state and \( z_t \) is a stochastic latent variable, $q_\phi(z_t | h_t, o_t) = \mathcal{N}\big(z_t; \mu_\phi(h_t, o_t), \sigma_\phi(h_t, o_t)\big)$, where \( \mu_\phi, \sigma_\phi \) are neural networks. 2) Sequence model: predicts the sequence of these representations given past actions $a_{t-1}$, $h_t = f_\theta(h_{t-1}, z_{t-1}, a_{t-1})$. 3) Dynamics predictor, predicts the prior latent state transition, $p_\theta(\hat{z}_t | h_t) = \mathcal{N}\big(\hat{z}_t; \mu_\theta(h_t), \sigma_\theta(h_t)\big)$. 4) Decoder: reconstructs observations from latent states, $p_\theta(o_t | h_t, z_t) = \mathcal{N}\big(o_t; \mu_\theta^{\text{obs}}(h_t, z_t), \sigma_\theta^{\text{obs}}\big)$. 5) Reward predictor, predicts rewards, $\hat{r}_t = r_\theta(h_t, z_t)$. 6) Continual predictor, predicts episode continuation flags, $\hat{c}_t = \text{sigmoid}\big(c_\theta(h_t, z_t)\big)$. Above all, RSSM can be defined as follows:
\begin{align}
  & \text{Encoder:} && z_t \sim q_\phi(z_t | h_t,o_t) \\
  & \text{Sequence model:} && h_t = f_\theta(h_{t-1},z_{t-1},a_{t-1}) \\
  & \text{Dynamics predictor:} && \hat{z}_t \sim p_\theta(\hat{z}_t | h_t) \\
  & \text{Decoder:} && \hat{o}_t \sim p_\theta(\hat{o}_t | h_t,z_t)\\
  & \text{Reward predictor:} && \hat{r}_t \sim r_\theta(\hat{r}_t | h_t,z_t) \\
  & \text{Continual predictor:} && \hat{c}_t \sim c_\theta(\hat{c}_t | h_t,z_t)
\end{align}

\section{EvolvingAgent}
\label{sec:EvolvingAgent}

Let $\mathcal{E}$ denote a dynamic open-world environment with partial observability, $\mathcal{T}$ represent the LH tasks, and $\mathcal{S}$ represents the agent's current state. We aim to design a curriculum self-evolving agent {\it EvolvingAgent} that can complete various LH tasks across environments, without human intervention. As shown in Fig. \ref{fig_2}, EvolvingAgent includes an experience-driven task planner $\Psi_{\text{plan}}$, a WM-guided action controller $\Pi_{\text{act}}$, a Curriculum Learning (CL) -based reflector $\Phi_{\text{reflect}}$; The continual world model includes a Multimodal Experience Pool (MEP) $\mathcal{D}_{\text{MEP}}$, and a world model $\mathcal{M}_w$. EvolvingAgent has an online MBRL setup and can be instantiated as:
\begin{equation}
\text{\it EvolvingAgent}: \langle \Psi_{\text{plan}}, \Pi_{\text{act}}, \Phi_{\text{reflect}}, \mathcal{D}_{\text{MEP}}, \mathcal{M}_w \rangle
\end{equation}

{\bf Continual world model}. As shown in Algorithm \ref{alg:evoagent}, by building a planner-controller-reflector closed-loop dynamic, the continual world model for EvolvingAgent can autonomously update multimodal experiences and world knowledge. The sketch is as follows:

\begin{equation}
\begin{matrix}
    \mathcal{E}, \mathcal{T}, \mathcal{S}, \mathcal{D}_{\text{MEP}}, \mathcal{M}_w \\
    \overbrace{\underbrace{\text{\bf Planner}}_{\substack{\Psi_{\text{plan}} \triangleright \mathcal{D}_{\text{MEP}} \\ \downarrow \{g_i\}}} \rightarrow
    \underbrace{\text{\bf Controller}}_{\substack{\Pi_{\text{act}} \circ \mathcal{M}_w \\ \downarrow {\{a_t\}, \mathcal{D}_{\text{MEP}}'}}} \rightarrow
    \underbrace{\text{\bf Reflector}}_{\substack{\Phi_{\text{reflect}} \triangleright \mathcal{D}_{\text{MEP}} \\ \downarrow \mathcal{M}_w^{'}}}\rightarrow}
\end{matrix}
\end{equation} 
where $\{g_i\}$ are subtasks generated by the planner $\Psi_{\text{plan}}$; $\{a_t\}$ are actions generated by the controller $\Pi_{\text{act}}$; $\mathcal{D}_{MEP}^{'}$ and $\mathcal{M}_w^{'}$ are the updated states.

{\bf Evaluation.} According to relevant research \cite{hafner2025mastering, Guo2024CaStLCA}, the agents' performance metrics includes Success Rate ({\bf SR}) and Exploration Efficiency ({\bf EE}).
\begin{equation}
    \text{SR}= \frac{\text{ENum}_{g_i}^\text{suc}}{\text{ENum}^{\text{all}}}, 
    \text{EE} = \frac{\text{Step}_{g_i}^{\text{suc}}}{\text{Step}_{g_i}^{\text{all}}} 
    \label{evaluation}
\end{equation}
where $\text{ENum}_{g_i}^\text{suc}$ indicates the number of episodes (a lifecycle) in which the subtask $g_i$ succeeded; $\text{ENum}_{g_i}^\text{all}$ indicates the total number of episodes; $\text{Step}_{g_i}^{\text{suc}}$ indicates the success step length of subtask $g_i$, and $\text{Step}_{g_i}^{\text{all}}$ indicates the total step length of subtask $g_i$ exploration. 

\subsection{Experience-driven Task Planner}
The experience-driven task planner $\Psi_{\text{plan}}$ is formalized as a function that maps the current state $\mathcal{S}$, LH task $\mathcal{T}$, and experience $\mathcal{D}_\text{MEP}$ to a sequence of subtasks $\mathcal{G}$.
\begin{equation}
    \Psi_{\text{plan}}: \mathcal{S} \times \mathcal{T} \times \mathcal{D}_{\text{MEP}} \rightarrow \mathcal{G}
\end{equation}
\begin{equation}
    \mathcal{S} = \mathcal{O}_{\text{obs}} \times \mathcal{S}_{\text{self}} \times \mathcal{S}_{\text{assets}},  s_t\in \mathcal{S}
\end{equation}
\begin{equation}
\label{eq12}
\mathcal{D}_{\text{MEP}} = \{h\}, h = \langle (s_t, a_t, r_t, s_{t+1}), \mathbb{P}_{(g_i)} | g_i \rangle
\end{equation}

where $\mathcal{G} = \{g_i\}_{i=1}^n$ is the subtask space, each subtask $g_i$ satisfies $\bigcup_{i=1}^n g_i \supseteq \mathcal{T}$; $\mathcal{O}_{\text{obs}}$ represents first-person observations, $\mathcal{S}_{\text{self}}$ represents the agent’s self-state, such as health or hunger, and $\mathcal{S}_{\text{assets}}$ represents agent's asset library, such as tools; $s_t$ represents the current state at step $t$; $h$ represents the experience; $r_t$ represents the reward obtained by performing action $a_t$ at state $s_t$; $\mathbb{P}(g_i)$ indicates the percentage of subtask $g_i$ completion.

{\bf Subtask planning.} As shown in Fig. \ref{fig_2}, we adopt the image tokenizer $f_{v}$ to encode the raw images $\mathcal{O}_{obs}, \mathcal{S}_{\text{self}}, \mathcal{S}_{\text{assets}}$ into token embeddings $\mathcal{V}=\{v_1,v_2,...,v_n\} \in \mathbb{R}^{n \times d}$, where $n$ denotes the number of visual tokens and $d$ is the dimensionality of each token. We adopt the textual tokenizer $f_{t}$ to encode $\mathcal{T}$ into token embeddings. We further utilize a lightweight projection module with a trainable projection matrix $W$. This module maps the visual tokens to the same space with text embeddings $\hat{\mathcal{V}} = W \mathcal{V}$, yielding $\hat{\mathcal{V}}=\{\hat{v}_1,\hat{v}_2,...,\hat{v}_n\} \in \mathbb{R}^{n \times d}$. The output of our planner is the subtask $g_i$.

{\bf Self-planning.} EvolvingAgent updates the planner for efficient LH task planning without human intervention. The LLM-based planner undergoes lightweight fine-tuning using Low-Rank Adaptation (LoRA)~\cite{hu2022lora}. The process of self-planning is as follows: 1) During agent initialization, the fine-tuning process utilizes all accumulated experiences from the multimodal experience pool $\mathcal{D}_{\text{MEP}}$ for task planning. When the multimodal experience pool is empty, the agent initializes task planning based on the capabilities of the original GPT-4o. 2) During the agent's lifecycle, when the feedback of WM-guided action controller indicates the subtask $g_i$ failure, experience trajectories relevant to the subtask $g_i$ by label matching are extracted to construct input-output pairs $\{(X_{\text{in}}^{(k)}, Y_{\text{out}}^{(k)})\}$ for model fine-tuning, where the input $X_{\text{in}}^{(k)}$ includes all the experience $h$ related the subtask $g_i$, while the output $Y_{\text{out}}^{(k)}$ represents whether the subtask in each experience was successful. We only use input-output pairs for model fine-tuning. This enables the planner to quickly study from the failure patterns while preserving its general planning capabilities, thereby improving robustness and reducing repeated errors in LH tasks. 3) When the agent dies (such as health value is 0), the agent is reinitialized.

\subsection{WM-Guided Action Controller}
The WM-guided action controller $\Pi_{\text{act}}$ is formalized as a function that maps the current state $\mathcal{S}$, subtask $\mathcal{G}$, and the world model $\mathcal{M}_w$ to an action sequence $a_{t:t+H} = \{a_t, a_{t+1}, \dots, a_{t+H}\}$ for horizon $H$.
\begin{equation}
\Pi_{\text{act}}: \mathcal{S} \times \mathcal{G} \times \mathcal{M}_w \rightarrow \mathcal{A}
\end{equation}

{\bf Action selection.} The controller utilizes $\mathcal{M}_w$ to predict future states and optimize actions:
\begin{equation}
    a_{t:t+H} = \underset{a_{t:t+H} \in \mathcal{A}^H}{\arg\max} \mathbb{E}_{\mathcal{M}_w} \left[ \sum_{\tau=t}^{t+H} \gamma^{\tau-t} R(s_\tau, a_\tau, g_i) \right]
\end{equation}
where $R(s_\tau, a_\tau, g_i)$ is the goal-aligned reward function, and $\gamma \in [0,1]$ is the discount factor. $R(s_\tau, a_\tau, g_i)$ considers not only the deterministic latent state and stochastic latent variable based on the current observation states $s_\tau$ and actions $a_\tau$ but also a goal embedding derived from the current subtask $g_i$, which is an extension of the DreamerV3~\cite{hafner2025mastering} reward. At each time step $t$, we sample a population of $N$ action sequences $\{a_{t:t+H}^{(k)}\}_{k=1}^{N}$ from the action space $\mathcal{A}^H$. The world model $\mathcal{M}_{w}$ is used to predict the future states and compute the expected cumulative reward $\mathbb{E}_{\mathcal{M}_{w}}\left[\sum_{\tau=t}^{t+H} \gamma^{\tau-t} R(s_{\tau}, a_{\tau}, g_i)\right]$ for each sequence. The sequence $a_{t:t+H}^*$ with the highest rewards is selected, and the first action $a_t^*$ of this sequence is executed in the environment.

{\bf Self-control.} EvolvingAgent uses a self-verification mechanism to reduce inefficient exploration and achieve efficient task execution without human intervention. After executing $a_t$, EvolvingAgent interacts with the environment $\mathcal{E}$ to collect environment feedback. Then EvolvingAgent uses a self-verification mechanism to determine whether the subtask $g_i$ can be terminated. The self-verification mechanism is as follows:
\begin{equation}
    \phi(s_t, g_i, t) = \begin{cases}
        \text{1} & \text{if } \cos(\text{Emb}_{s_t}, \text{Emb}_{g_i}) \geq \sigma \lor t \geq T_{\text{max}} \\
        \text{0} & \text{otherwise}
    \end{cases}
\end{equation}
where $\phi(s_t, g_i, t)=1$ indicates that the subtask $g_i$ can be terminated; $\text{Emb}_{s_t}$ is the WM-encoded latent representation of the current state $s_t$, $\text{Emb}_{g_i}$ is the task embedding derived from the subtask $g_i$ description. Similar to MINEDOJO \cite{fan2022minedojo}, we trained a contrastive video-language model pre-trained on the multimodal experience pool. It computes the cosine similarity $\cos(\cdot)$ between an open-vocabulary language goal embedding $\text{Emb}_{g_i}$ and an 8-frame video snippet embedding $\text{Emb}_{s_t}$, which is used to measure goal attainment with threshold $\sigma$ set empirically. $T_{\text{max}}$ is the maximum allowed steps of each episode. When a subtask $g_i$ is completed or the subtask $g_i$ completion cycle exceeds the maximum step length $T_{\text{max}}$, the subtask $g_i$ is terminated and the experience-driven task planner is performed again.

If the subtask $g_i$ is terminated, whether it is successful or exceeds the step threshold, $\left\{ \langle s_t, a_t, r_t, s_{t+1}, \mathbb{P}_{(g_i)} | g_i \rangle \right\}_{t}$ is added to the multimodal experience pool $\mathcal{D}_{\text{MEP}}$.
\begin{equation}
    \mathcal{D}_{\text{MEP}^{'}} \leftarrow \mathcal{D}_{\text{MEP}} \cup \left\{ \langle s_t, a_t, r_t, s_{t+1}, \mathbb{P}_{(g_i)} | g_i \rangle \right\}_{t}
\end{equation}

\subsection{CL-based Reflector}
The CL-based reflector $\Phi_{\text{reflect}}$ is formalized as a function that maps the current state $\mathcal{S}$, subtask $\mathcal{G}$, and the multimodal experience $\mathcal{D}_{\text{MEP}}$ to update the world model from $\mathcal{M}_w$ to $\mathcal{M}'_w$.
\begin{equation}
\Phi_{\text{reflect}}: \mathcal{S} \times \mathcal{G} \times \mathcal{D}_{\text{MEP}} \times \mathcal{M}_w \rightarrow \mathcal{M}'_w
\end{equation}

$\Phi_{\text{reflect}}$ employs a two-stage CL algorithm to optimize experience selection, which can enable agents to efficiently update the world model without human intervention as the agent interacts dynamically with the environment.

{\bf Stage 1: curriculum subtask selection.} For candidate subtasks $g_i \in \mathcal{G}$, we use four indicators for curriculum subtask selection: (1) the relevance of the subtask $g_i$ to the current target task $\mathcal{T}_{goal}$; (2) the exploration efficiency of the subtask $g_i$ (the ratio of successful step length $\text{Step}_{g_i}^{\text{suc}}$ to total step length $\text{Step}_{g_i}^{\text{all}}$); (3) the importance of the subtask $g_i$ (comparing its impact on the current world model $\mathcal{M}_{w,g_i}^{\text{new}}$ and past world model $\mathcal{M}_{w,g_i}^{\text{old}}$); (4) the average completion ratio $\overline{\mathbb{P}}_{(g_i)}$ of the subtask $g_i$.

Therefore, the priority score $\tau(g_i)$ of the subtask $g_i$ can be defined as follows:
\begin{equation}
\begin{aligned}
    \tau(g_i) = &\underbrace{\lambda_1 \cdot \cos(\text{Emb}_{g_i}, \text{Emb}_{\mathcal{T}_{goal}})}_{\text{Relevance}} + \underbrace{\lambda_2 \cdot \frac{\text{Step}_{g_i}^{\text{suc}}}{\text{Step}_{g_i}^{\text{all}}}}_{\text{Efficiency}}\\& + \underbrace{\lambda_3 \cdot \text{KL}\left(\mathcal{M}_{w,g_i}^{\text{old}} \| \mathcal{M}_{w,g_i}^{\text{new}}\right)}_{\text{Importance}} + \underbrace{\lambda_4 \cdot \overline{\mathbb{P}}_{g_i}}_{\text{Completion Ratio}}
    \label{eq:task_score}
\end{aligned}
\end{equation}
where $\cos(\text{Emb}_{g_i}, \text{Emb}_{\mathcal{T}_{goal}})$ represents the cosine similarity of task embedding. $\lambda_1+\lambda_2+\lambda_3 +\lambda_4=1$ are balancing coefficients. Finally, in round $k$, $|\mathcal{D}_k^{\text{subtask}}|$ subtasks are selected. 
\begin{equation}
    \mathcal{D}_k^{\text{subtask}} = \{g_i | \tau(g_i) \geq \rho_k \}, \quad 
    \rho_k = \rho_0 \cdot e^{-c_s k}
\end{equation}
where $c_s$ controls the curriculum subtask progression rate.

{\bf Stage 2: curriculum experience selection.} For candidate experience $h \in \mathcal{D}_{MEP}$ in selected subtasks $\mathcal{D}_k^{\text{subtask}}$, we use three indicators for curriculum experience selection: (1) the Temporal Difference Error (TD-Error) $\delta_{\text{TD}}(h_j)$, prioritizes experience with high TD-Error, indicating prediction mismatch between current and target world models; (2) the Gradient Norm $\| \nabla_{\mathcal{M}_w} \mathcal{L}_{\text{pred}}(h_j) \|$, favors experiences that maximally influence the world model's parameter updates; (3) the Information Gain, measures how much the experience $h_j$ changes the world model's belief distribution, calculated via KL divergence between current $\mathcal{M}_w^{\text{new}}(s_{j+1}|h_j)$ and previous $\mathcal{M}_w^{\text{old}}(s_{j+1}|h_j)$ world model predictions.
\begin{equation}
\begin{aligned}
    \epsilon(h_j) = &\eta_1 \cdot \underbrace{|\delta_{\text{TD}}(h_j)|}_{\text{TD-Error}} + \eta_2 \cdot \underbrace{\|\nabla_{\mathcal{M}_w} \mathcal{L}_{\text{pred}}(h_j)\|_2}_{\text{Gradient Norm}} \\ & +\underbrace{\eta_3 \cdot \text{KL}\left(\mathcal{M}_w^{\text{new}}(s_{j+1}|h_j) \| \mathcal{M}_w^{\text{old}}(s_{j+1}|h_j)\right)}_{\text{Information Gain}}
\end{aligned}
\end{equation}
where $\eta_1+\eta_2+\eta_3=1$ are balancing coefficients. Finally, in round $k$, $|\mathcal{D}_k^{\text{exp}}|$ experiences are selected. 
\begin{equation}
    \mathcal{D}_k^{\text{exp}} = \{h_j | \epsilon(h_j) \geq \rho_k \}, \quad 
    \rho_k = \rho_0 \cdot e^{-c_h k}
\end{equation}
where $c_h$ controls curriculum experience progression rate.

\input{table/alg_1}

\textbf{Self-reflection.} Update the world model $\mathcal{M}_w$ using experiences $\mathcal{D}_k^{\text{exp}}$ with importance-aware weight $w_j$:
\begin{equation}
    \mathcal{M}_w^{'} \leftarrow \mathcal{M}_w - \nabla \Big[ \underbrace{\textstyle{\sum}_{h_j} w_j \mathcal{L}_{\text{pred}}(h_j)}_{\text{Curriculum Loss}} + \underbrace{\mu \cdot \Omega(\theta, \theta^{\text{old}})}_{\text{Regularization}} \Big]
\end{equation}
\begin{equation}
    w_j = \frac{\epsilon(h_j)}{\max_k \epsilon(h_k)}, \Omega = \sum_i \mathcal{F}_i (\theta_i - \theta_i^{\text{old}})^2
\end{equation}
where $w_j$ to emphasize critical experiences, and $\Omega$ to penalize shifts in parameters critical for past tasks. $\mathcal{F}_i$ is the Fisher information matrix diagonal.

\section{Experiments}
In this section, we conduct experiments on Minecraft to verify the performance of EvolvingAgent. We also include detailed ablation studies to analyze the effectiveness of each component, sensitivity analysis to evaluate the contribution of each indicator, and generalization analysis to test the generalization ability.

\subsection{Experimental Setting}
{\bf Simulators}. We use Minecraft \cite{fan2022minedojo} to evaluate EvolvingAgent. Minecraft features a procedurally generated 3D world of different biomes, which consists of 1-meter-sized blocks that the player can break and place. There are about 30 different creatures that the player can interact with or fight. We employ MineRL 0.4.4 with Minecraft as our simulation environment. The agent operates at a fixed speed of 20 frames per second and only interacts with the environment via low-level control signals. Optimus-1~\cite{li2024optimus1hybridmultimodalmemory} constructs a benchmark of 67 tasks to evaluate the Agent’s ability for LH tasks. We use the same task group partitioning as the Optimus-1 to evaluate EvolvingAgent. We also evaluate the cross-environment generalization of our method on Atari100k. Atari100k \cite{bellemare2013arcade} is a standard reinforcement learning benchmark based on the Arcade Learning Environment, covering a diverse set of Atari 2600 games under a limited interaction steps.

\input{table/table_1}

\input{table/table_2}

{\bf Hyperparameters}. EvolvingAgent is designed based on the codebase of dreamerV3 \cite{hafner2025mastering}. The planner of EvolvingAgent uses the VQ-GAN \cite{esser2021taming} and GPT-4o for task planning. The controller of EvolvingAgent is based on the RSSM-based WM structure~\cite{hafner2025mastering} for action selection. For detailed hyperparameters, please refer to the Appendix \ref{hyper}. Among them, about the self-verification threshold $\sigma$, we perform a sensitivity analysis by running our agent in Minecraft with $10^7$ environment steps for the task "Iron". as shown in Table \ref{SA_1}, experimental results show that the task success rate remains stable when $\sigma \in [0.875, 0.925]$, with sharp declines outside this range due to over/under-termination. When $\sigma \textless 0.875$, sub-tasks may not be completed but are misjudged, causing subsequent tasks to fail. When $\sigma \textgreater 0.925$, due to strict self-verification, sub-tasks may be completed but still require re-planning, reducing the task completion rate.

{\bf Training.} EvolvingAgent runs on a single A100 GPU. Taking $10^7$ steps as an example, compared to dreamerV3 running for 7 days, EvolvingAgent only needs to run for 2.7 days.

\textbf{LLM API Call}. LLM API calls occur in two processes: subtask planning and planner fine-tuning. As the agent self-evolves, the number of subtask failures decreases, which greatly reduces the overhead of planner fine-tuning. Throughout the experiment, with an average of 750 planning calls over $10^7$ environment steps.

{\bf Baselines.} We compare EvolvingAgent with existing outperforming agents, including model-free Agent (PPO \cite{schulman2017proximalpolicyoptimizationalgorithms}), WM-based agents (dreamerV3 \cite{hafner2025mastering}, LS-Imagine \cite{li2025open}) and LLM-based agents (GPT-4V, Jarvis-1 \cite{Wang2023JARVIS1OM}, Optimus-1 \cite{li2024optimus1hybridmultimodalmemory}) on the challenging LH tasks cross-environments. We do not consider agents that are completely based on human data and curricula support (such as Voyager \cite{wang2023voyager}, DEPS \cite{wang2023describe}, and Steve-Eye \cite{zheng2023steve}).

\subsection{ Quantitative Results and Analysis}
As shown in Table \ref{table1}, EvolvingAgent achieves state-of-the-art success rates (SR) and exploration efficiency (EE) across all resource tiers. Compared with existing methods, EvolvingAgent can achieve an average success rate improvement of {\bf 111.74{\%}} and reduce ineffective actions by more than {\bf 6x}. 

For basic tasks (Wood/Stone), EvolvingAgent marginally outperforms Optimus-1 (97.47{\%} vs. 96.39{\%} SR on Wood) but exhibits significantly greater advantages in advanced tasks like Gold (21.69{\%} vs. 10.62{\%} SR) and Diamond (17.36{\%} vs. 9.30{\%} SR). This hierarchy-aligned improvement suggests EvolvingAgent’s planner-controller-reflector closed-loop dynamic effectively addresses LH dependencies, where traditional model-based methods (DreamerV3 and LS-Imagine) and LLM-driven agents (Jarvis-1) struggle to maintain coherent multi-stage strategies. Notably, the EE reveals EvolvingAgent’s exploration superiority: its 30.48{\%} EE on Gold tasks is 3.8× higher than Optimus-1, indicating reduced invalid actions.

Model-free methods (PPO) and pure vision-language models (GPT-4V) fail completely (0{\%} SR/EE) on tasks requiring tool hierarchies (Iron+), highlighting their inability to model latent state transitions. While Jarvis-1 and DreamerV3 achieve partial success on intermediate tasks (42.38{\%} SR on Iron), their performance collapses on Gold/Diamond tiers due to compounding errors in action sequences. LS-Imagine, through a hybrid approach of short-term and long-term imagination, significantly outperforms DreamerV3 in SR and EE metrics on the basic Wood/Stone task. However, its performance growth is slow on complex tasks because accumulated errors can mislead the optimization direction. EvolvingAgent with 26.83{\%} EE on Diamond tasks, 7.3× higher than Optimus-1, underscores how CL-based experience selection mitigates exploration bottlenecks in sparse-reward scenarios.

\input{table/table_3}

\subsection{Ablation Studies}
The ablation study reveals critical insights into the contributions of each components (Planner without LoRA, Planner with LoRA, Controller, Reflector only with stage 1, Reflector only with stage 2, Reflector with both stages) and Continual WM for LH tasks. We selected 10 random seeds for testing. Table \ref{tb2} shows the mean and variance of the average success rate (SR) for each ablation study. When only PPO is used without any modules (first row), the agent fails to progress beyond basic tasks (28.16{\%} SR for Wood, 0{\%} for Iron+). Introducing the Planner module nearly doubles performance on Wood (45.69{\%}) and marginally improves Stone (18.37{\%}), but still fails to unlock advanced tasks (Iron+ at 0{\%}), suggesting that planner alone cannot resolve the exploration bottleneck in LH tasks. A pivotal leap occurs when Controller is added (P+C), with Wood and Stone success rates surging to 92.42{\%} and 85.31{\%}, respectively, and modest progress in Iron (31.59{\%}). This underscores the necessity of structured exploration to navigate intermediate dependencies. However, the sharp decline in Gold (5.47{\%}) and Diamond (3.52{\%}) indicates persistent challenges in sparse reward scenarios. Integrating the Reflector module (P+C+R) achieves near-perfect Wood/Stone success (96.69{\%}/93.82{\%}) and significantly boosts Iron (42.61{\%}), Gold (17.53{\%}), and Diamond (10.09{\%}), demonstrating its role in distilling exploration experiences to refine world models.

\begin{figure}[t]
\centering
	\includegraphics[width=0.48\textwidth]{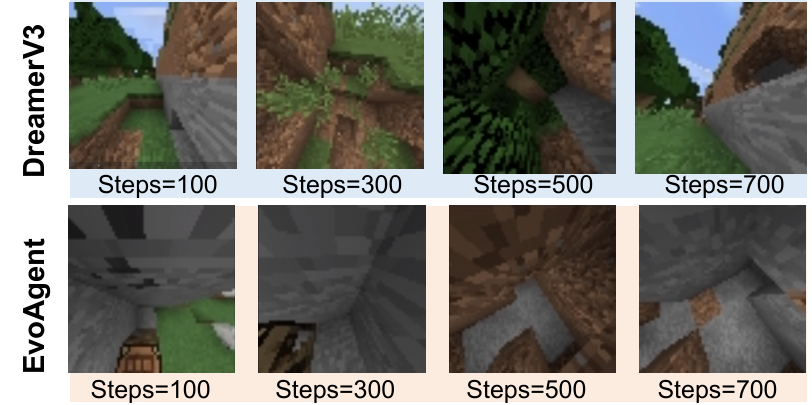}
	\centering
	\caption{ Illustration of the role of CL-based reflector. Take the subtask "Craft a stone axe" as an example.}
	\label{fig_3}
\end{figure}

This experiment compares the results of Planner with LoRA and Planner without LoRA, demonstrating that planner fine-tuning can significantly improve model convergence speed, reduce the number of invalid subtasks, and achieve autonomous task decomposition. We compare the effects of Reflector with stage 1 only, Reflector with stage 2 only, and Reflector with both stages. The experimental results show that combining curriculum subtask selection with curriculum experience selection can significantly improve the model's average performance. EvolvingAgent is an improvement over DreamerV3. As shown in Fig. \ref{fig_3}, EvolvingAgent is significantly better than DreamerV3 in path selection, which can complete subtasks in the fewest steps. This is because CL-based reflectors can efficiently update the world model through subtask selection and experience selection, reducing the impact of redundant experience on the world model update. EvolvingAgent with the CL-based reflector can greatly reduce invalid exploration and accelerate task completion.

\input{table/table_4}

\subsection{Sensitivity Analysis}
We further examine the effect of different indicators in curriculum subtask selection, curriculum experience selection, and self-reflection on the \textit{Iron} task group. As shown in Table \ref{table4}, adding more indicators consistently improves the success rate in all three parts. For curriculum subtask selection, introducing relevance, efficiency, importance, and completion ratio gradually raises the success rate from 40.16{\%} to 49.37{\%}, showing that effective subtask prioritization requires not only task similarity but also execution feedback and world-model variation. For curriculum experience selection, combining TD-Error, gradient norm, and information gain improves the success rate from 42.47{\%} to 50.43{\%}, indicating that informative experiences should be selected from prediction discrepancy, update influence, and knowledge increment simultaneously. For self-reflection, adding regularization to curriculum loss further improves the success rate from 48.61{\%} to 50.92{\%}, suggesting that stable world-model updating is important for reliable long-horizon task completion.

\subsection{Generalization Analysis}
We further evaluate the cross-environment generalization of EvolvingAgent on Atari100k after training in Minecraft. As shown in Table \ref{table5}, EvolvingAgent consistently outperforms DreamerV3 under the same 400K interaction steps on most Atari tasks, including Alien, Assault, Asterix, Battle Zone, Ms Pacman, and Road Runner. It also reaches or exceeds human-level performance on several games such as Boxing, Krull, and Kung Fu Master. These results indicate that the continual world model learned by EvolvingAgent captures transferable dynamics and control priors rather than environment-specific patterns only. This cross-domain advantage supports the robustness and scalability of EvolvingAgent under substantial environment shifts.

\input{table/table_5}

\section{Conclusion}
This paper presents EvolvingAgent, an embodied agent that improves long-horizon task execution through curriculum self-evolution. EvolvingAgent integrates an experience-driven task planner, a WM-guided action controller, and a CL-based reflector to continuously refine subtask scheduling, experience selection, and world model updating during interaction. This design enables more reliable planning and execution in open-ended environments. In the future, we hope that our method can be truly applied to real robot scenarios.

\bibliographystyle{IEEEtran}
\bibliography{./IEEEabrv, ./TMM26} 

\newpage
\onecolumn
\appendix
\label{hyper}
All hyperparameters of EvolvingAgent are shown in the Table \ref{table6}.
\input{table/table_6}

\end{document}

%% file: table/alg_1.tex
\begin{algorithm}[t]
\caption{Continual World Model}
\label{alg:evoagent}
\begin{algorithmic}[1]
\REQUIRE Environment $\mathcal{E}$, LH task $\mathcal{T}$, current state $ \mathcal{S}$, MEP $\mathcal{D}_{\text{MEP}}$, world model $\mathcal{M}_w$, horizon $H$, max steps $T_{\text{max}}$
\ENSURE Updated $\mathcal{D}_{\text{MEP}}^{'}$, $\mathcal{M}_w^{'}$

\FOR{LH Task $\mathcal{T}=\mathcal{T}_0$ to $\mathcal{T}_n$}
        \STATE {\textcolor{blue!60!black}{Experience-driven task planner via Eq. (13-15)}}
        \STATE $\{g_i\} \gets \Psi_{\text{plan}}(\mathcal{S}, \mathcal{T}, \mathcal{D}_{\text{MEP}})$
    \FOR{each subtask ${g_i} \in \{g_i\}$} 
        \FOR{episode $t=1$ to $T_{\text{max}}$}
            \STATE {\textcolor{blue!60!black}{WM-guided action controller via Eq. (16-19)}}
            \STATE $\{a_{t:t+H}\} \gets \Pi_{\text{act}}(s_t, g_i, \mathcal{M}_w)$
            \IF{$\phi(s_t, g_i, t)$ is Terminal}
            \STATE $\mathcal{D}_{\text{MEP}}^{'} \gets \mathcal{D}_{\text{MEP}} \cup \{\langle s_t,a_t,r_t,s_{t+1},\mathbb{P}_{(g_i)}|g_i \rangle\}$
            \STATE BREAK
            \ENDIF
        \ENDFOR
        \STATE {\textcolor{blue!60!black}{CL-based reflector via Eq. (20-26)}}
        \STATE $\mathcal{D}_k^{\text{subtask}} \gets \text{Curriculum\_Subtask\_Select}(\mathcal{G}_t, \mathcal{T}, \mathcal{D}_{\text{MEP}})$
        \STATE $\mathcal{D}_k^{\text{exp}} \gets \text{Curriculum\_Experience\_Select}(\mathcal{D}_k^{\text{subtask}})$
        \STATE $\mathcal{M}_w^{'} \gets \Phi_{\text{reflect}}
        (\mathcal{D}_k^{\text{exp}}, \mathcal{M}_w)$
    \ENDFOR
\ENDFOR
\end{algorithmic}
\end{algorithm}

%% file: table/table_1.tex
\begin{table*}[t]
\centering
\small
\caption{The results of all the methods on the Minecraft. The evaluation metrics are the average success rate (SR) and the average exploration efficiency (EE) (as shown in Eq. \ref{evaluation}). Upper EE metrics mean that the agent is more efficient at completing the task with fewer invalid exploration steps, while $0.00$ indicates that the agent is unable to complete the task. The Overall represents the average result on the three groups of Iron \includegraphics[width=0.25cm]{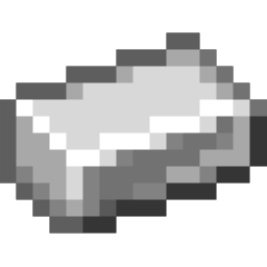}, Gold \includegraphics[width=0.25cm]{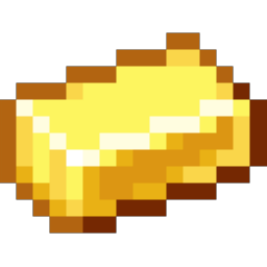}, and Diamond \includegraphics[width=0.25cm]{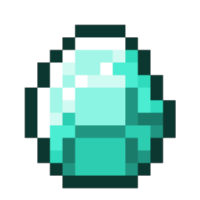}. The Improving represents the average performance improvement of EvoAgent compared to the algorithms Jarvis-1, dreamerV3, LS-Imagine, and Optimus-1. The best results are in bold.}
\label{table1}
\resizebox{\textwidth}{!}{%
\renewcommand\arraystretch{1.25}
\begin{tabular}{c|ccccccc>{\columncolor[HTML]{ECF4FF}}c>{\columncolor[HTML]{ECF4FF}}c} 
\toprule
{\bf Group} & {\bf Metric}       & {\bf PPO}   & {\bf GPT-4V} & {\bf Jarvis-1} & {\bf DreamerV3} & {\bf LS-Imagine} & {\bf Optimus-1} & {\bf EvoAgent} & {\bf Improving} ({\%})\\
\midrule
\multirow{2}{*}{\includegraphics[width=0.4cm]{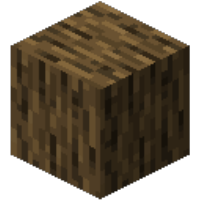} Wood}         & SR$\uparrow$ & 28.16 & 35.24  & 89.73    & 91.07   & 95.87  & 96.39     & {\bf 97.47}  & 4.51 $\uparrow$  \\
 & EE$\uparrow$ & 53.82 & 69.45  & 87.36    & 93.22    & 97.41 & 97.82     & {\bf 98.43} & 4.77 $\uparrow$ \\ \midrule
\multirow{2}{*}{\includegraphics[width=0.4cm]{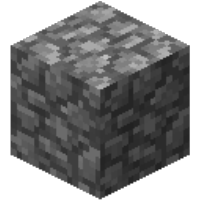} Stone} & SR$\uparrow$ & 13.42 & 14.39  & 81.91    & 86.82   & 91.50  & 88.79  & {\bf 94.53}  & 8.34 $\uparrow$  \\
& EE$\uparrow$ & 27.56 & 30.64  & 84.72    & 88.39   & 92.36  & 89.25     & {\bf 96.48} & 8.80 $\uparrow$  \\ \midrule
\multirow{2}{*}{\includegraphics[width=0.4cm]{picture/iron_ingot.pdf} Iron}  & SR$\uparrow$ & 0.00  & 0.00   & 42.38    & 33.79  & 35.82   & 45.48     & {\bf 51.82}  & 3.16 $\uparrow$  \\
& EE$\uparrow$ & 0.00  & 0.00   & 47.52    & 35.68  & 38.27   & 46.16     & {\bf 58.54} & 39.67 $\uparrow$   \\ \midrule
\multirow{2}{*}{\includegraphics[width=0.4cm]{picture/gold_ingot.pdf} Gold}  & SR$\uparrow$ & 0.00  & 0.00   & 8.84     & 6.57    & 6.61  & 10.62      & {\bf 21.69}    & {\bf {\color{red} 165.81 $\uparrow$}}\\
& EE$\uparrow$ & 0.00  & 0.00   & 9.76     & 8.05   & 10.69   & 8.03      & {\bf 30.48} &233.75 $\uparrow$   \\ \midrule
\multirow{2}{*}{\includegraphics[width=0.4cm]{picture/diamond.pdf} Diamond}   & SR$\uparrow$ & 0.00  & 0.00   & 7.69     & 4.73   & 4.36   & 9.30      & {\bf 17.36}    & 166.26 $\uparrow$\\
& EE$\uparrow$ & 0.00  & 0.00   & 0.07     & 3.69   & 4.19   & 7.31      & {\bf 26.83}    & {\bf {\color{red} 603.28 $\uparrow$}}\\ \midrule
\rowcolor[HTML]{ECF4FF}
Overall & SR$\uparrow$ & 0.00  & 0.00   & 19.64    & 15.03  & 15.60   & 21.80     & {\bf 30.29} & {\bf {\color{red} 111.74 $\uparrow$}}  \\
\bottomrule[1.3pt] 
\end{tabular}
}
\end{table*}

%% file: table/table_2.tex
\begin{table}[h]
\centering
\caption{The sensitivity analysis of the hyperparameter $\sigma$.}
\label{SA_1}
\resizebox{0.5\textwidth}{!}{%
\begin{tabular}{c|ccc>{\columncolor[HTML]{ECF4FF}}ccc}
\toprule
$\sigma$       &0.825   & 0.85  & 0.875 & 0.9   & 0.925 & 0.95 \\ \midrule
SR & 17.03 & 34.48 & 48.69 & 52.43 & 47.72 & 26.74 \\
\bottomrule
\end{tabular}
}
\end{table}

%% file: table/table_3.tex
\begin{table*}[t] 
\centering
\caption{Ablation study results. We report the average success rate (SR) on each task group. \texttt{P.$^{-}$}, \texttt{P.}, \texttt{C.}, \texttt{R.$^{1}$}, \texttt{R.$^{2}$}, \texttt{R.}, and \texttt{CWM} represent Planner without LoRA, Planner with LoRA, Controllor, Reflector only with stage 1, Reflector only with stage 2, Reflector with both stages, and Continual World Model, respectively. The PPO algorithm is used by default for model decision-making.  Numbers after the$\pm$ signs represent standard deviations. The best results are in bold.}
\resizebox{\textwidth}{!}{%
\renewcommand\arraystretch{1.25}
\begin{tabular}{>{\columncolor[HTML]{ECF4FF}}ccc>{\columncolor[HTML]{ECF4FF}}c>{\columncolor[HTML]{ECF4FF}}ccc|ccccc}
\toprule
\multicolumn{7}{c|}{{\bf Setting}}                                                                                   & \multicolumn{5}{c}{{\bf Tasks}}                 \\\midrule
P.$^{-}$   &  P.   & C.   & R.$^{1}$  & R.$^{2}$ & R.  & CWM  & Wood  & Stone  & Iron  & Gold  & Diamond \\ \midrule
 & &  &  &  & &   & 28.16$_{\pm 6.01}$ & 13.42$_{\pm 7.62}$ & 0.00$_{\pm 0.00}$ & 0.00$_{\pm 0.00}$ & 0.00$_{\pm 0.00}$ \\
\Checkmark && &  &   &  &  & 41.36$_{\pm 5.20}$ & 16.27$_{\pm 8.32}$ & 0.00$_{\pm 0.00}$ & 0.00$_{\pm 0.00}$ & 0.00$_{\pm 0.00}$ \\
& \Checkmark &  &  &   &   &   & 45.69$_{\pm 5.12}$ & 18.37$_{\pm 6.71}$ & 0.00$_{\pm 0.00}$ & 0.00$_{\pm 0.00}$ & 0.00$_{\pm 0.00}$ \\
& \Checkmark & \Checkmark &   &    &    &    & 92.42$_{\pm 3.31}$ & 85.31$_{\pm 5.96}$ & 31.59$_{\pm 6.72}$ & 5.47$_{\pm 2.46}$ & 3.52$_{\pm 2.31}$ \\
& \Checkmark & \Checkmark & \Checkmark &  &   &   & 93.18$_{\pm 2.86}$ & 87.72$_{\pm 5.74}$ & 34.63$_{\pm 5.19}$ & 8.68$_{\pm 2.72}$ & 5.14$_{\pm 2.51}$ \\
& \Checkmark & \Checkmark & & \Checkmark &  & & 95.37$_{\pm 2.48}$ & 91.26$_{\pm 3.86}$ & 39.58$_{\pm 5.08}$ & 14.20$_{\pm 4.05}$ & 8.93$_{\pm 3.73}$ \\
& \Checkmark & \Checkmark &  & & \Checkmark &  & 96.69$_{\pm 2.24}$ & 93.82$_{\pm 3.34}$ & 42.61$_{\pm 4.80}$ & 17.53$_{\pm 5.18}$ & 10.09$_{\pm 3.54}$ \\ \midrule
\rowcolor[HTML]{ECF4FF} & \Checkmark & \Checkmark & &  & \Checkmark & \Checkmark & {\bf 97.47}$_{\pm 1.75}$ & {\bf 94.53}$_{\pm 2.82}$ & {\bf 51.82}$_{\pm 4.60}$ & {\bf 21.69}$_{\pm 4.61}$ & {\bf 17.36}$_{\pm 2.34}$ \\
\bottomrule
\end{tabular}
}
\label{tb2}
\end{table*}

%% file: table/table_4.tex
\begin{table}[t]
\centering
\caption{The sensitivity analysis about indicators in the curriculum subtask selection (Eq. 21): Relevance (R.), Efficiency (E.), Importance (I.), and Completion Rate (C.R.); indicators in the curriculum experience selection (Eq. 23): TD-Error (TD-R.), Gradient Norm (G.N.), and Information Gain (I.G.); indicators in the self-reflection (Eq. 25): Curriculum Loss (C.L.) and Regularization (R.). We report the average success rate on the task "Iron" group.}
\renewcommand\arraystretch{1}
\begin{tabular}{>{\centering\arraybackslash}p{1.5cm}|>{\centering\arraybackslash}p{0.8cm}>{\centering\arraybackslash}p{0.8cm}>{\centering\arraybackslash}p{0.8cm}>{\centering\arraybackslash}p{0.8cm}|>{\centering\arraybackslash}p{1.3cm}}
\toprule
{\bf Equation}                     & \multicolumn{4}{c|}{{\bf Indicators}} & {\bf SR}    \\ \midrule
\multicolumn{1}{c|}{}    & R.   & E.     & I.    & C.R.  & Iron  \\ \midrule
\multirow{4}{*}{\textbf{Equation 21}} & \Checkmark    &        &       &       & 40.16 \\
                        & \Checkmark    & \Checkmark      &       &       & 41.77 \\
                        & \Checkmark    & \Checkmark      & \Checkmark     &       & 46.54 \\
                         & \cellcolor[HTML]{ECF4FF}\Checkmark    & \cellcolor[HTML]{ECF4FF}\Checkmark      & \cellcolor[HTML]{ECF4FF}\Checkmark     & \cellcolor[HTML]{ECF4FF}\Checkmark     & \cellcolor[HTML]{ECF4FF}{\bf 49.37} \\ \midrule
                        &      & TD-R.  & G.N.  & I.G.  &   Iron    \\ \midrule
\multirow{3}{*}{\textbf{Equation 23}} &      & \Checkmark      &       &       & 42.47 \\ 
                        &      & \Checkmark      & \Checkmark     &       & 45.59 \\
                        &   \cellcolor[HTML]{ECF4FF}   & \cellcolor[HTML]{ECF4FF} \Checkmark      & \cellcolor[HTML]{ECF4FF} \Checkmark     & \cellcolor[HTML]{ECF4FF} \Checkmark     & \cellcolor[HTML]{ECF4FF} {\bf 50.43} \\ \bottomrule
                        &      &        & C.L.  & R.    &    Iron   \\ \midrule
\multirow{2}{*}{\textbf{Equation 25}} &      &        & \Checkmark     &       & 48.61 \\
                         &   \cellcolor[HTML]{ECF4FF}   &   \cellcolor[HTML]{ECF4FF}     & \cellcolor[HTML]{ECF4FF}\Checkmark     & \cellcolor[HTML]{ECF4FF}\Checkmark     & \cellcolor[HTML]{ECF4FF}{\bf 50.92}\\
                        \bottomrule
\end{tabular}
\label{table4}
\end{table}

%% file: table/table_5.tex
\begin{table}[t]
\centering
\small
\caption{The generalization analysis in Atari100k.}
\label{table5}
\renewcommand\arraystretch{1.2}
\resizebox{0.5\textwidth}{!}{%
\begin{tabular}{c|c|cc|>{\columncolor[HTML]{ECF4FF}}c}
\toprule
\textbf{Task} & \textbf{Human} & \textbf{PPO} & \textbf{DreamerV3} & \textbf{EvoAgent} \\
\midrule
Steps &  — & 400K & 400K & 400K \\
\midrule
Alien  & 7128 & 276 & 1118 & \textbf{1392} \\
Amidar  & 1720 & 26 & 97 & \textbf{329}\\
Assault  & 742 & 327 & 683 & \textbf{981}\\
Asterix  & 8503 & 292& 1062 & \textbf{1492}\\
Bank Heist  & 753 & 14& \textbf{398} & 362\\
Battle Zone  & 37188 & 2233 & 20300 & \textbf{24830}\\
Boxing  & 12 & 3 & 82 & \textbf{91}\\
Breakout & 30 & 3 & 10 & \textbf{13}\\
Chopper Command  & 7388 & 1005 &2222 & \textbf{4375}\\
Crazy Climber  & 35829 & 14675 & \textbf{86225} & 78215\\
Demon Attack  & 1971 & 160 & 577 & \textbf{1205}\\
Freeway & 30 & 2 & 0 & \textbf{5}\\
Frostbite  & 4335 & 127 & 3377 & \textbf{3674}\\
Gopher  & 2412 & 368 & 2160 & \textbf{2219}\\
Hero  & 30826 & 2596 & \textbf{13354} & 12168\\
Jamesbond & 303 & 41 & 540 & \textbf{621}\\
Kangaroo  & 3035 & 55& 2643 & \textbf{2753}\\
Krull  & 2666 & 3222 &8171 & \textbf{10027}\\
Kung Fu Master & 22736 & 2090 & 25900 & \textbf{28692}\\
Ms Pacman & 6952 & 366& 1521 & \textbf{3246}\\
Pong  & 15 & -20 & -4 & \textbf{-2}\\
Private Eye & 69571 & 100 & 3238 & \textbf{5285}\\
Qbert & 13544 & 317 & 2921 & \textbf{4793}\\
Road Runner & 7845 & 602 & 19230 & \textbf{21703}\\
Seaquest & 42055 & 305 & 962 & \textbf{2305}\\
Up N Down & 11693 & 1502 & \textbf{46910} & 37284\\
\bottomrule
\end{tabular}
}
\end{table}

%% file: table/table_6.tex
\begin{table}[h]
\centering
\caption{EvoAgent hyperparameters.}
\label{table6}
\resizebox{0.66\textwidth}{!}{
\begin{tabular}{|l|c|c|}
\toprule
\multicolumn{3}{|c|}{\textbf{General}} \\
\midrule
Replay capacity & — & \(5 \times 10^6\) \\
Batch size & \(B\) & 16 \\
Batch length & \(T\) & 64 \\
Activation & — & RMSNorm + SiLU \\
Learning rate & — & \(4 \times 10^{-5}\) \\
Gradient clipping & — & AGC(0.3) \\
Optimizer & — & LaProp(\(\epsilon = 10^{-20}\)) \\
\midrule
\multicolumn{3}{|c|}{\textbf{World Model}} \\
\midrule
Reconstruction loss scale & \(\beta_{\text{pred}}\) & 1 \\
Dynamics loss scale & \(\beta_{\text{dyn}}\) & 1 \\
Representation loss scale & \(\beta_{\text{rep}}\) & 0.1 \\
Latent unimix & — & 1\% \\
Free nats & — & 1 \\
\midrule
\multicolumn{3}{|c|}{\textbf{Actor Critic}} \\
\midrule
Imagination horizon & \(H\) & 15 \\
Discount horizon & \(1/(1 - \gamma)\) & 333 \\
Return lambda & \(\lambda\) & 0.95 \\
Critic loss scale & \(\beta_{\text{val}}\) & 1 \\
Critic replay loss scale & \(\beta_{\text{repval}}\) & 0.3 \\
Critic EMA regularizer & — & 1 \\
Critic EMA decay & — & 0.98 \\
Actor loss scale & \(\beta_{\text{pol}}\) & 1 \\
Actor entropy regularizer & \(\eta\) & \(3 \times 10^{-4}\) \\
Actor unimix & — & 1\% \\
Actor RetNorm scale & \(S\) & \(\text{Per}(R, 95) - \text{Per}(R, 5)\) \\
Actor RetNorm limit & \(L\) & 1 \\
Actor RetNorm decay & — & 0.99 \\
\midrule
\multicolumn{3}{|c|}{\textbf{ WM-Guided Action Controller}} \\
\midrule
Maximum episode step length & $T_{max}$ & 24000 \\
Task similarity threshold & $\sigma$ & 0.9 \\
Reward discount factor & $\gamma$ & 0.1 \\
\midrule
\multicolumn{3}{|c|}{\textbf{ CL-based Reflector}} \\
\midrule
CL algorithm initialization threshold & $\rho_0$ & $5 \times 10^{-3}$ \\
CL subtask selection increase rate & $c_s$ & 0.3 \\
CL experience selection increase rate & $c_h$ & 0.5 \\
World model penalize weight & $\mu$ & 0.1\\
\bottomrule
\end{tabular}
}
\end{table}